\documentclass[conference,10pt]{IEEEtran}
\IEEEoverridecommandlockouts
\usepackage{amsmath,amssymb,amsfonts}
\usepackage{adjustbox}
\usepackage{algorithmic}
\usepackage{threeparttable}
\usepackage{array}
\usepackage{fontawesome}
\usepackage{graphicx}
\usepackage{multirow}
\usepackage{textcomp}
\usepackage{enumitem}
\usepackage[english]{babel}
\usepackage[utf8]{inputenc}
\usepackage[nolist]{acronym}
\usepackage{amsmath}											
\usepackage{amsfonts}											
\usepackage{blindtext} 											
\usepackage{booktabs}											
\usepackage{fancyvrb}
\usepackage{ifthen}	
\usepackage{wrapfig}											
\usepackage{filecontents} 										
\usepackage[T1]{fontenc} 
\usepackage{graphicx}
\usepackage[dvipsnames,table]{xcolor}
\usepackage{ifthen}												
\usepackage{listings}											
\usepackage{lipsum}
\usepackage{multirow}
\usepackage{pgfplots}
\usepackage{siunitx}
\usepackage{ragged2e}						
\usepackage{tabularx}
\usepackage{tikz}
\usepackage{circuitikz}
\usepackage{comment}
\usepackage{footnote}
\usepackage{float}
\usepackage{balance}
\usepackage{setspace}
\usepackage{makecell, tabularx}

\usetikzlibrary{calc,trees,positioning,arrows,chains,shapes.geometric,%
	decorations.pathreplacing,decorations.pathmorphing,shapes,%
	matrix,shapes.symbols}

\usepackage[font=footnotesize,labelfont=bf]{subcaption}		
\usepackage{caption}
\usepackage[ruled,linesnumbered]{algorithm2e}
\SetAlCapNameFnt{\small}
\SetAlCapFnt{\small}
\SetAlgoNlRelativeSize{0}
\RestyleAlgo{ruled}
\SetKwComment{Comment}{/* }{ */}

\usepackage[style=ieee, backend=biber, natbib=true, maxbibnames=1, minbibnames=1]{biblatex}
\usepackage{fancyhdr}

\fancypagestyle{firstpage}{%
	\fancyhf{}%
	\fancyhead[C]{\normalsize To appear at the IEEE International Conference on IC Design and Technology 2026 (ICICDT), \\June 22 - 24, 2026, Dresden, Germany}
	
	\fancyfoot[C]{\footnotesize \copyright2026 IEEE. Personal use of this material is permitted. Permission from IEEE must be obtained for all other uses, in any current or future media, including reprinting/republishing this material for advertising or promotional purposes, creating new collective works, for resale or redistribution to servers or lists, or reuse of any copyrighted component of this work in other works.}

}

\lstdefinelanguage{Verilog}{morekeywords={accept_on,alias,always,always_comb,always_ff,always_latch,and,assert,assign,assume,automatic,before,begin,bind,bins,binsof,bit,break,buf,bufif0,bufif1,byte,case,casex,casez,cell,chandle,checker,class,clocking,cmos,config,const,constraint,context,continue,cover,covergroup,coverpoint,cross,deassign,default,defparam,design,disable,dist,do,edge,else,end,endcase,endchecker,endclass,endclocking,endconfig,endfunction,endgenerate,endgroup,endinterface,endmodule,endpackage,endprimitive,endprogram,endproperty,endspecify,endsequence,endtable,endtask,enum,event,eventually,expect,export,extends,extern,final,first_match,for,force,foreach,forever,fork,forkjoin,function,generate,genvar,global,highz0,highz1,if,iff,ifnone,ignore_bins,illegal_bins,implements,implies,import,incdir,include,initial,inout,input,inside,instance,int,integer,interconnect,interface,intersect,join,join_any,join_none,large,let,liblist,library,local,localparam,logic,longint,macromodule,matches,medium,modport,module,nand,negedge,nettype,new,nexttime,nmos,nor,noshowcancelled,not,notif0,notif1,null,or,output,package,packed,parameter,pmos,posedge,primitive,priority,program,property,protected,pull0,pull1,pulldown,pullup,pulsestyle_ondetect,pulsestyle_onevent,pure,rand,randc,randcase,randsequence,rcmos,real,realtime,ref,reg,reject_on,release,repeat,restrict,return,rnmos,rpmos,rtran,rtranif0,rtranif1,s_always,s_eventually,s_nexttime,s_until,s_until_with,scalared,sequence,shortint,shortreal,showcancelled,signed,small,soft,solve,specify,specparam,static,string,strong,strong0,strong1,struct,super,supply0,supply1,sync_accept_on,sync_reject_on,table,tagged,task,this,throughout,time,timeprecision,timeunit,tran,tranif0,tranif1,tri,tri0,tri1,triand,trior,trireg,type,typedef,union,unique,unique0,unsigned,until,until_with,untyped,use,uwire,var,vectored,virtual,void,wait,wait_order,wand,weak,weak0,weak1,while,wildcard,wire,with,within,wor,xnor,xor,`uvm_create, `uvm_rand_send_with},morecomment=[l]{//}}

\newboolean{blindreview}
\setboolean{blindreview}{false}
\DeclareRobustCommand{\IEEEauthorrefmark}[1]{\smash{\textsuperscript{\footnotesize #1}}}

\usepackage[left=1.5cm,right=1.5cm,top=2cm,bottom=3cm]{geometry}

\begin{document}	
\begin{acronym}[] 
	\acro{BST}[BST]{Binary Search Tree}
	\acro{CEX}[CEX]{Counterexample}
	\acrodefplural{CEX}[CEXs]{Counterexamples}
	\acro{FV}[FV]{Formal Verification}
	\acro{IR}[IR]{Intermediate Representation}
	\acrodefplural{IR}[IRs]{Intermediate Representations}
	\acro{KG}[KG]{Knowledge Graph}
	\acrodefplural{KG}[KGs]{Knowledge Graphs}
	\acro{LLM}[LLM]{Large Language Model}
	\acrodefplural{LLM}[LLMs]{Large Language Models}
	\acro{RAG}[RAG]{Retrieval Augmented Generation}
	\acro{RTL}[RTL]{Register Transfer Level}
	\acro{SV}[SV]{SystemVerilog}
	\acro{SVA}[SVA]{SystemVerilog Assertion}
	\acrodefplural{SVA}[SVAs]{SystemVerilog Assertions}
\end{acronym}


\title{Knowledge Graphs, the Missing Link in Agentic AI-based Formal Verification

\thanks{Under grant 101194371, Rigoletto is supported by Chips Joint Undertaking and its members, including top-up funding by the National Funding Authorities from involved countries. Rigoletto is also funded by the Federal Ministry of Research, Technology and Space under the funding code 16MEE0548S. The responsibility for the content of this publication lies with the author.}
}

\ifthenelse{\boolean{blindreview}}
{}{
	\author{
		\IEEEauthorblockN{
			Vaisakh Naduvodi Viswambharan\IEEEauthorrefmark{1},
			Keerthan Kopparam Radhakrishna\IEEEauthorrefmark{1},
			Deepak Narayan Gadde\IEEEauthorrefmark{1},
			Aman Kumar\IEEEauthorrefmark{2}
			}
		\IEEEauthorblockA{
			\IEEEauthorrefmark{1}Infineon Technologies Dresden AG \& Co. KG, Germany\\
			\IEEEauthorrefmark{2}Infineon Technologies Semiconductor India Private Limited, India
			}
		}
}

\maketitle
\thispagestyle{firstpage}

\begin{abstract}

Recent advances in \acp{LLM} have enabled workflows that generate \acp{SVA} from natural-language specifications, with the potential to accelerate \ac{FV}. However, high-quality assertion synthesis remains challenging because specifications are often ambiguous or incomplete and critical micro-architectural details reside in the \ac{RTL}. Many existing approaches treat the specification and \ac{RTL} as loosely structured text, which weakens specification-to-\ac{RTL} grounding and leads to semantic mismatches and frequent syntax failures during formal parsing and elaboration. This work addresses these limitations with a verification-centric \ac{KG} constructed from structured \acp{IR} extracted from the specification, \ac{RTL}, and formal-tool feedback, including syntax diagnostics, \acp{CEX}, and coverage reports. The \ac{KG} links requirements, design hierarchy, signals, assumptions, and properties to provide traceable, design-grounded context for generation. A multi-agent workflow queries and updates this \ac{KG} to generate \acp{SVA} and to drive three refinement loops: syntax repair guided by tool diagnostics, \ac{CEX}-guided correction using trace links, and coverage-directed property augmentation. Evaluation across seven benchmark designs indicates that \ac{KG}-based context retrieval improves specification-to-\ac{RTL} grounding and consistently produces compilable \acp{SVA} with low syntax-repair overhead. The approach achieves formal coverage ranging from 78.5\% to 99.4\%, though convergence exhibits design dependence with complex temporal and arithmetic reasoning remaining challenging for current \ac{LLM} capabilities.

\end{abstract}


\section{Introduction} \label{sec:introduction}
\ac{FV} offers strong correctness for \ac{RTL} designs by proving that key behaviors hold across all reachable states, including corner cases that are difficult to cover with simulation. Central to \ac{FV} is using \acp{SVA} to encode intended behavior as formal properties. However, writing accurate assertions requires deep understanding of both the specification and  \ac{RTL} implementation, making assertions time-consuming and error-prone.

Advances in \acp{LLM} enable automated synthesis of \acp{SVA} from natural-language specifications \cite{10458667,10.1145/3658617.3697756,10682649,kumar2025saarthiaiformalverification,11218681}. 
To produce semantically correct \acp{SVA}, models need context linking to the relevant \ac{RTL} hierarchy, and signal interactions \cite{11105590}. 
Accordingly, prior work builds context through structured specification processing, staged prompting \cite{10.1145/3658617.3697756,wu2025spec2assertionautomaticprertlassertion} and \ac{RTL}-derived prompt enrichment \cite{10740198,lyu2025assertminermodulelevelspecgeneration}. 
Common strategies include schema-guided normalization \cite{10682649} and retrieval-augmented context selection \cite{10830777,11130226}.
However, studies report that these assertions show syntax and semantic errors requiring iterative correction and manual triage \cite{10682649,10992817}, motivating representations with explicit, persistent links across requirements, \ac{RTL} semantics, generated properties, and verification outcomes.

\ac{KG}-based structuring offers more explicit grounding. AssertionForge \cite{11105590} builds a \ac{KG} linking specification artifacts to the \ac{RTL} hierarchy and signals for structured context  Yet, formal-tool outcomes, including \acp{CEX}, and coverage gaps, are typically not represented as  queryable \ac{KG} content, limiting traceability and hindering systematic closed-loop refinement.

To address these limitations, this work proposes a verification-centric methodology that builds a \ac{KG} from structured \acp{IR} extracted from the specification, \ac{RTL}, and formal-tool feedback. \ac{LLM} agents query and update the \ac{KG} to support design-grounded property generation and three refinement loops: syntax repair guided by parser and elaboration diagnostics, \ac{CEX}-guided correction using traces, and coverage-directed augmentation from tool-reported gaps.

The main contributions of this paper are as follows:
\begin{itemize}
	\item A verification-centric \ac{KG} construction based on structured \acp{IR}, integrating specification intent, \ac{RTL} structure, generated properties, and formal-tool results.
	\item A multi-agent workflow that uses \ac{KG}-grounded context to generate and refine \acp{SVA} with traceability across requirements, design elements, properties, and results.
	\item An evaluation on open-source designs quantifying the impact of \ac{KG}-grounded context on property generation effectiveness, syntactic validity, and \ac{CEX} refinement.
\end{itemize}

\section{Related Work} \label{sec:background}
\begin{figure*}[htbp!]
	\centering
	
	\begin{subfigure}{\textwidth}
		\centering
		\includegraphics[width=\textwidth,height=0.38\textheight,keepaspectratio]{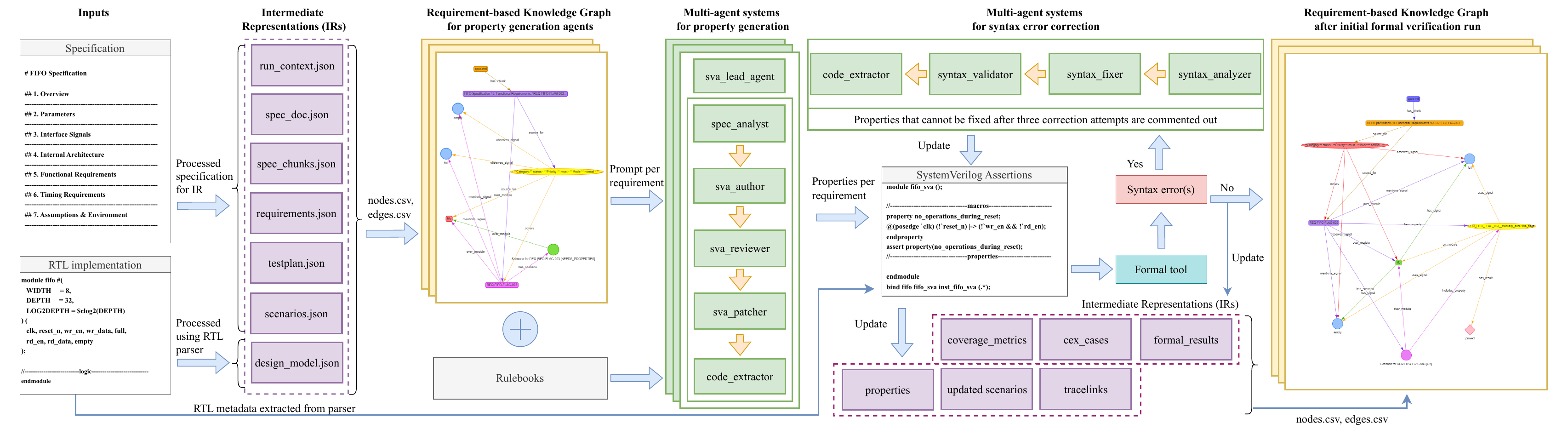}
		\caption{Knowledge Graphs and multi-agent systems used for property generation and syntax error correction}
		\label{fig:workflow_part1}
	\end{subfigure}
	
	\vspace{0.6em}
	
	\begin{subfigure}{\textwidth}
		\centering
		\includegraphics[width=\textwidth,height=0.38\textheight,keepaspectratio]{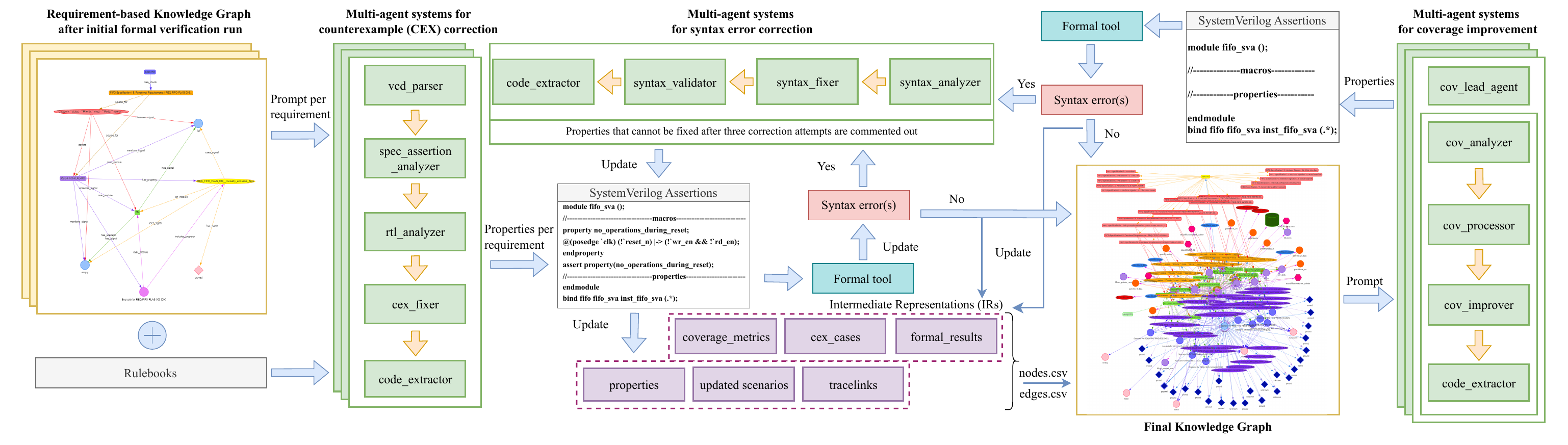}
		\caption{Knowledge Graphs and multi-agent systems used for CEX correction, coverage improvement and syntax error correction}
		\label{fig:workflow_part2}
	\end{subfigure}
	
	\caption{End-to-end workflow of the proposed approach. NetworkX is used for KG construction, and PyVis provides interactive HTML visualization.}
	\label{fig:workflow_overview}
\end{figure*}

In \ac{LLM}-assisted \ac{SVA} generation, prior works differ in representing specification intent and design context; key challenge is grounding requirements and \ac{RTL} semantics, while enabling iterative refinement from formal verification outcomes.

Specification-driven approaches synthesize \acp{SVA} from natural-language specifications, but are limited by ambiguous and missing behavioral constraints \cite{10458667,10.1145/3658617.3697756}. To improve reliability, many systems structure specifications before generation: ChIRAAG \cite{10682649} uses JSON schema-guided representations, AssertLLM \cite{10.1145/3658617.3697756} decomposes extraction and synthesis while Spec2Assertion \cite{wu2025spec2assertionautomaticprertlassertion} improves functional correctness via progressive regularization and structured reasoning yet remaining largely pre-\ac{RTL}. These approaches improve consistency but struggle when micro-architectural details are absent.

In contrast, \ac{RTL}-driven approaches use static analysis of implementation artifacts to guide assertion generation, improving signal relevance but risking intent drift without explicit code links \cite{10740198,lyu2025assertminermodulelevelspecgeneration}. Hybrid approaches connect intent and implementation more explicitly: AssertionForge \cite{11105590} constructs a \ac{KG} linking specification artifacts to the \ac{RTL} hierarchy and signals for structured context. Retrieval augmentation and  structured prompting further focus models on relevant context: \ac{RAG}-based approaches retrieve targeted fragments to reduce hallucinations \cite{10830777}, and LISA \cite{11130226} combines retrieval with structured reasoning to improve temporal correctness and reduce retries. While effective, these methods rely on complete cross-artifact links between requirements, signals, and behaviors.

Despite these advances, studies still report recurring syntax errors and semantic mismatches requiring iterative fixes \cite{10992817}. Some systems add limited feedback loops, such as simulation- or test-driven repair prompting \cite{10682649,10651860}. Saarthi \cite{kumar2025saarthiaiformalverification} advances with an agentic, end-to-end formal workflow coordinating verification planning, assertion generation, formal proving, and coverage evaluation, yet formal-tool outputs remain transient rather than persistent, queryable artifacts.

Overall, prior work improves \ac{SVA} generation via structured prompting, retrieval, and agentic decomposition but traceability remains weak without structured,  queryable tool outputs. Thus, this work introduces a verification- centric \ac{KG} constructed from structured \acp{IR} spanning specification intent, \ac{RTL} structure, generated properties, and formal-tool feedback. This foundation enables multi-agent generation and closed-loop refinement driven by syntax diagnostics, \acp{CEX}, and coverage gaps, with traceability from requirements to assertions and  formal outcomes.

\section{Methodology} \label{sec:methodology}

\begin{table*}[t]
	\centering
	\small
	\setlength{\tabcolsep}{6pt}
	\renewcommand{\arraystretch}{1.08}
	\caption{Intermediate Representation artifacts and their roles in the verification workflow}
	\begin{tabular}{p{0.14\textwidth} p{0.86\textwidth}}
		\hline
		\textbf{IR Artifact} & \textbf{Purpose} \\
		\hline
		
		\textit{spec\_chunks.json} &
		Hierarchical specification decomposition with heading structure, semantic tags, enabling targeted context retrieval \\
		\hline
		
		\textit{requirements.json} &
		Extracted requirements with identifiers, natural language text, categories, priorities, traceability to specification chunks \\
		\hline
		
		\textit{testplan.json} &
		Per-requirement verification plans, observable signals, stimulus-response behaviors, timing constraints \\
		\hline
		
		\textit{design\_model.json} &
		RTL metadata from formal tool analysis: module hierarchies, ports (direction, bit-width), signals, FSM encodings, parameters \\
		\hline
		
		\textit{properties.json} &
		Generated \ac{SVA} code with property identifiers, requirement traceability, type classification, line mappings \\
		\hline
		
		\textit{tracelinks.json} &
		Mappings encoding requirement-to-specification, property-to-requirement, property-to-coverage relationships for bidirectional traceability \\
		\hline
		
		\textit{formal\_results.json} &
		Per-property outcomes with formal status (proven, CEX, vacuous), proof depth, runtime, diagnostic artifacts (VCD) \\
		\hline
		
		\textit{cex\_cases.json} &
		Counterexample tracking with property identifiers, VCD paths, failure line numbers, correction attempt history for iterative debugging \\
		\hline
		
		\textit{coverage\_metrics.json} &
		Aggregated statistics: reachability percentages, proof core ratio, vacuity counts, dead code classifications driving coverage iterations \\
		\hline
		
		\textit{run\_context.json} &
		Execution snapshot with run identifiers, file paths, iteration counts, tool versions, timestamps, configuration \\
		\hline
		
		\textit{nodes.csv} &
		Graph node export with unique IDs, type labels (spec\_chunk, requirement, property, formal\_result), run IDs, JSON attributes \\
		\hline
		
		\textit{edges.csv} &
		Graph edge export with source/destination IDs, relationship types (derives\_from, validates, proves, fails, covers), run IDs, JSON attributes \\
		\hline
	\end{tabular}
	\label{tab:ir_artifacts}
\end{table*}

The workflow is organized into four layers:
\begin{enumerate}[label=(\roman*), leftmargin=*]
	\item Inputs: specification, \ac{RTL}, configuration, and agents
	\item Typed JSON \acp{IR}: normalize requirements, \ac{RTL} structure, properties, and verification outcomes
	\item Runtime \acp{KG}: retrieves localized context neighborhoods bounded by task relevance and provides traceable, design-grounded context
	\item Agent orchestration: routes property generation, syntax/\ac{CEX} repair, and coverage improvement loops while preserving traceability from requirements to properties and outcomes
\end{enumerate}

In each run, requirements and \ac{RTL} metadata are extracted into typed JSON \acp{IR}, followed by property generation and formal verification. Syntax and \ac{CEX} correction along with coverage improvement are triggered based on tool outputs and updated \ac{IR} artifacts as shown in Fig.~\ref{fig:workflow_overview}.

\begin{lstlisting}[
	basicstyle=\scriptsize\ttfamily,
	frame=single,
	caption={IR core entity types and relationships},
	label={IRSchemaListing}
	]
	Entities:
	SpecChunk        - specification fragments
	Requirement      - functional requirements
	Property         - generated SVA code
	FormalResult     - verification outcomes
	CoverageMetrics  - RTL coverage tracking
	Relationships:
	Requirement     --> SpecChunk (provenance)
	Property        --> Requirement (validation)
	Property        --> FormalResult (outcomes)
	Property/RTL    --> CoverageMetrics (coverage)
\end{lstlisting}

\subsection{Intermediate Representations and Knowledge Graphs}
\label{sec:ir_kg}

Verification artifacts are stored as typed JSON \acp{IR} to enable deterministic orchestration and cross-iteration comparison. Table~\ref{tab:ir_artifacts} enumerates the \ac{IR} artifacts, and Listing~\ref{IRSchemaListing} summarizes the core entity types and trace links used for requirement-centric, property-centric, and coverage-centric retrieval.

The runtime \ac{KG} is constructed from \texttt{nodes.csv} and \texttt{edges.csv} exported from the \ac{IR} artifacts. The \ac{KG} acts as an artifact registry with property-granular updates: when a property is regenerated or corrected, only its downstream evidence (for example, formal results, \acp{CEX}, and coverage records) is invalidated and recomputed, enabling focused re-verification while preserving end-to-end traceability.

The runtime \ac{KG} constructs bounded context by traversing from a task anchor node, typically a requirement or a property. For property generation, the neighborhood includes the target requirement, its parent specification fragment, relevant \ac{RTL} objects (module scope and signals), and related properties validating sibling requirements. For repair tasks, the neighborhood also includes tool outputs such as compilation diagnostics, \ac{CEX} traces, and coverage gap reports. \ac{RTL} metadata is indexed by hierarchical signal identifiers, enabling automatic mapping from specification-level signal mentions to \ac{RTL} hierarchical paths.

\subsection{Multi-Agent Orchestration}
\label{sec:multiagent}

Verification is decomposed into specialized agents to separate concerns that would be difficult for a single monolithic prompt: specification interpretation, \ac{RTL} signal grounding, temporal logic synthesis, and tool-driven debugging. Each agent operates with a \ac{KG}-bundled context neighborhood appropriate to its task, emits structured updates to \ac{IR} objects and property code, and hands off to the next agent through explicit coordination protocols. This enables deterministic routing and complete auditability of the verification workflow. Fig.~\ref{fig:workflow_part1} illustrates the agent interactions during property generation, while Fig.~\ref{fig:workflow_part2} shows the \ac{CEX} correction and coverage improvement systems.

\subsubsection{Property Generation Agents}
\label{sec:pipe_gen}

Requirements are translated into \ac{SVA} properties through a multi-agent workflow. The \texttt{sva\_lead} oversees the property generation process by coordinating agents and ensuring verification strategy alignment with requirements. \ac{RTL} metadata extracted from the parser populates the \texttt{design\_model} \ac{IR} artifact with module hierarchies, ports, signals, and parameters. For each requirement, the \ac{KG} provides a neighborhood containing the requirements, parent specification fragment, \ac{RTL} signals, and rulebook guidelines. The \texttt{spec\_analyst} decomposes requirements into testable conditions: trigger events, responses, timing constraints, and exceptions. The \texttt{sva\_author} translates these into \ac{SVA} syntax using assertions, temporal operators, and assume directives. The \texttt{sva\_reviewer} validates against specification adherence and rulebooks. The \texttt{sva\_patcher} corrects properties based on \texttt{sva\_reviewer} feedback. The \texttt{code\_extractor} assembles approved properties into a \ac{SVA} file with macros and property declarations.

\begin{table*}[tbp!]
	\centering
	\scriptsize
	\setlength{\tabcolsep}{6pt}
	\renewcommand{\arraystretch}{1.08}
	\caption{Evaluation of the proposed methodology}
	\begin{tabular}{p{0.13\textwidth} p{0.16\textwidth} c c c c c c c}
		\hline
		\multicolumn{1}{c}{\textbf{Evaluation Category}} &
		\multicolumn{1}{c}{\textbf{Evaluation Metric}} &
		\textbf{FIFO \cite{opencores_web}} &
		\textbf{Lemming \cite{pinckney2025comprehensiveverilogdesignproblems}} &
		\textbf{ALU \cite{pinckney2025comprehensiveverilogdesignproblems}} &
		\textbf{CICD \cite{pinckney2025comprehensiveverilogdesignproblems}} & 
		\textbf{BST \cite{pinckney2025comprehensiveverilogdesignproblems}} & \textbf{TTC Counter \cite{pinckney2025comprehensiveverilogdesignproblems}} & 
		\textbf{CSR \cite{pinckney2025comprehensiveverilogdesignproblems}} \\
		\hline
		\textit{} & \#Properties (T | \textcolor{green}{P} | \textcolor{red}{F})  
		& 26 | \textcolor{green}{17} | \textcolor{red}{9}
		& 51 | \textcolor{green}{50} | \textcolor{red}{1}
		& 2 | \textcolor{green}{1} | \textcolor{red}{1}
		& 15 | \textcolor{green}{12} | \textcolor{red}{3}
		& 34 | \textcolor{green}{18} | \textcolor{red}{16}
		& 41 | \textcolor{green}{30} | \textcolor{red}{11}
		& 42 | \textcolor{green}{26} | \textcolor{red}{16} \\
		\textit{Property generation} & \#Vacuous properties    
		& 0 & 0 & 0 & 0 & 7 & 2 & 5 \\
		\textit{} & \%Reachable RTL coverage                 
		& 88.2 & 94.7 & 95.8 & 72.6 & 73.1 & 59.0 & 33.8 \\
		\hline
		\textit{Syntax correction} & \#Fix attempts           
		& 1 & 0 & 3 & 1 & 1 & 1 & 1 \\
		\hline
		\textit{CEX correction} & \#Properties (C | NC)    
		& 8 | 1 & 0 | 1 & 0 | 1
		& 0 | 3 & 0 | 16 & 0 | 11 & 0 | 16  \\
		\hline
		\textit{} & \#Properties (T | \textcolor{green}{P} | \textcolor{red}{F})  
		& 26 | \textcolor{green}{25} | \textcolor{red}{1}
		& 51 | \textcolor{green}{50} | \textcolor{red}{1}
		& 2 | \textcolor{green}{1} | \textcolor{red}{1}
		& 15 | \textcolor{green}{12} | \textcolor{red}{3}
		& 37 | \textcolor{green}{20} | \textcolor{red}{17}
		& 41 | \textcolor{green}{30} | \textcolor{red}{11}
		& 42 | \textcolor{green}{32} | \textcolor{red}{10} \\
		\textit{Coverage improvement} & \#Vacuous properties   
		& 1 & 0 & 0 & 0 & 7 & 2 & 5 \\
		\textit{} & \%Reachable RTL coverage                 
		& 88.2 & 94.7 & 95.8 & 71.8 & 73.1 & 59.0 & 84.6 \\
		\hline
		\textit{End-to-end results} & \%Formal coverage       
		& 93.8 & 97.32 & 99.41 & 90.83 & 86.46 & 82.98 & 93.82 \\
		\hline
		\textit{Knowledge Graph size} & \#Nodes | \#Edges      
		& 231 | 1054 & 402 | 2157 & 116 | 396
		& 165 | 688 & 299 | 1107 & 302 | 1522 & 338 | 1397  \\
		\hline
	\end{tabular}%
	\label{tab:results}
	\vspace{1ex}
	\noindent
	\begin{minipage}{\textwidth}
		\vspace{1em}
		\scriptsize 
		T - Total, P - Passed, F - Failed, C - Corrected, NC - Not Corrected, FIFO - First In First Out, ALU - Arithmetic Logic Unit, CICD - Cascaded Integrator-Comb Decimator, BST - Binary Search Tree, TTC Counter - Time-to-Completion Counter, CSR (CSR APB Interface Design) - Control and Status Register Advanced Peripheral Bus Interface
	\end{minipage}
\end{table*}

\subsubsection{Syntax Correction Agents}
\label{sec:pipe_syntax}

\ac{LLM}-generated syntax errors in properties are addressed through a multi-agent correction strategy. After compilation, the \texttt{syntax\_analyzer} parses error messages and attributes them to specific properties. The \texttt{syntax\_fixer} applies deterministic repairs for well-known patterns (undeclared identifiers wrapped with backtick macro syntax or hierarchical paths, undefined macros auto-resolved by adding definitions) and handles complex errors (temporal operators, sequence expressions, type mismatches) using compilation diagnostics, property context, requirement text, and \ac{RTL} signal declarations. The \texttt{syntax\_validator} performs isolated testing in a temporary wrapper to prevent cascading errors before reintegration. The \texttt{code\_extractor} extracts the corrected code. Properties receive maximum three attempts; persistent failures are disabled and logged for manual review.

\subsubsection{Counterexample Correction Agents}
\label{sec:pipe_cex}

Properties failing \ac{FV} produce \acp{CEX} indicating \ac{RTL} bugs or property issues. VCD waveforms, failure timestamps, and signals are extracted and stored as FormalResult \ac{IR} entities. The \texttt{vcd\_parser} extracts temporal behavior at failure points. The \texttt{spec\_assertion\_analyzer} classifies root causes: \ac{RTL} bugs, over-specification, missing assumptions, or under-specification. The \texttt{rtl\_analyzer} examines \ac{RTL} source code and traces logic paths; genuine bugs are documented for \ac{RTL} correction. The \texttt{cex\_fixer} corrects property issues by adjusting constraints, adding assumptions, or strengthening assertions. The \texttt{code\_extractor} assembles corrected properties for isolated verification before reintegration. Properties receive maximum three attempts, with persistent failures flagged for manual analysis.

\subsubsection{Coverage Improvement Agents}
\label{sec:pipe_cov}

Syntactically correct properties may still achieve inadequate coverage. Formal tool reports are parsed to extract coverage metrics: unreachable \ac{RTL} statements, dead code regions and vacuous properties, stored as CoverageMetrics \ac{IR} entities. The \texttt{cov\_lead\_agent} coordinates coverage improvement by prioritizing gaps in functional paths and distinguishing intentionally unreachable defensive code. The \texttt{cov\_analyzer} performs root cause analysis, examining \ac{RTL} implementation to identify conditions for unreachable code and querying the \ac{KG} to determine if existing properties or assumptions prevent those combinations. The \texttt{cov\_processor} links gaps to requirements through \ac{KG} traversal, identifying under-verified requirement aspects. The \texttt{cov\_improver} generates targeted properties using cover directives for reachability and assertions for correctness. The \texttt{code\_extractor} assembles coverage properties for iterative formal runs.

\section{Evaluation} \label{sec:results}

The system was evaluated on seven benchmark designs: FIFO, Lemming, ALU, CIC Decimator, \ac{BST}, TTC Counter, and CSR APB Interface, using the multi-agent flow with the GPT-5.2 model and Cadence JasperGold for \ac{FV}, as shown in Table~\ref{tab:results}. Overall, the workflow consistently produces runnable \ac{SVA} at scale. Syntax correction remains low effort across the completed designs, which indicates that \ac{KG}-grounded context, especially deterministic signal and hierarchy resolution via the \texttt{design\_model} \ac{IR}, reduces common front-end issues such as undeclared identifiers, macro mismatches, and incorrect hierarchical references.

The main differences across designs appear after compilation, during \ac{CEX}-based refinement and coverage improvement. FIFO and CIC Decimator benefit from iterative refinement, with improved pass rates and higher end-to-end formal coverage after the correction loops. This suggests that many initial failures are caused by local issues such as missing guards, overly strong assumptions, or boundary-condition constraints, which can be corrected using retrieved context and \ac{CEX} evidence. TTC Counter shows moderate convergence: a non-trivial fraction of initially failing properties become provable after \ac{CEX} correction, but overall coverage remains lower than the strongest benchmarks, consistent with limited reachable-state coverage in the design. CSR APB Interface shows a different pattern: initial property quality is lower, but the coverage loop yields a clear improvement in passing properties, indicating that additional targeted properties help exercise previously uncovered behavior.

In contrast, Lemming and ALU show limited benefit from automated correction. Most properties pass early, and the remaining failures do not improve significantly, suggesting that the unresolved cases require complex temporal reasoning such as reset behavior, FSM transition intent, or multi-cycle arithmetic meaning. \ac{BST} remains the most difficult: many properties are generated, but a large fraction remain failing and \ac{CEX} correction does not converge, which is consistent with the need for global data-structure  invariants and inductive reasoning that exceed local patching strategies.

Vacuous-property behavior further separates the benchmarks. CIC Decimator and \ac{BST} exhibit more vacuous properties, indicating some checks can pass without strongly constraining behavior under the default environment. This motivates stronger environment modeling or non-vacuity constraints earlier in the loop so coverage improvements reflect exercised behavior rather than trivially satisfied properties.

\ac{KG} size grows with design complexity and trace link density, but it does not directly predict verification success. Overall, the \ac{KG} improves reliability and repeatability of artifact management and context retrieval, particularly for property generation and syntax stabilization, while semantic debugging remains the main limiting factor on harder benchmarks.

\section{Conclusion} \label{sec:conclusion}

This paper presented a \ac{KG}-centric, multi-agent workflow for \ac{LLM}-assisted \ac{FV} that addresses specification-to-\ac{RTL} grounding through structured \acp{IR} and bounded context retrieval. Specifications, \ac{RTL} metadata, properties, tool outcomes, and coverage evidence are represented as typed JSON artifacts and exported as nodes and edges for constructing a requirement-based runtime \ac{KG} that supports property generation and post-run refinement loops. Across the evaluated benchmarks, the approach consistently produces compilable, traceable \acp{SVA} with low syntax-fix overhead, while the largest remaining challenge is semantic correction for properties that require deeper temporal reasoning or global invariants. Importantly, as \acp{LLM} improve, they can better leverage the \ac{KG}’s structured context and traces to improve property generation and downstream formal verification. Future work will focus on improving \ac{CEX} correction with richer temporal reasoning patterns and inductive invariant generation for stateful designs, and on refining coverage-gap classification to better separate unreachable code from true verification targets.

\printbibliography  

\end{document}